\documentclass[10pt,twocolumn,letterpaper]{article}

\usepackage{wacv}
\usepackage{times}
\usepackage{epsfig}
\usepackage{graphicx}
\usepackage{amsmath}
\usepackage{amssymb}

\usepackage[T1]{fontenc}    
\usepackage{color}
\usepackage{multirow}

\usepackage{algorithm}
\usepackage[noend]{algpseudocode}

\newsavebox{\tempbox}

\wacvfinalcopy 

\ifwacvfinal
\def\assignedStartPage{1} 
\fi

\ifwacvfinal
\usepackage[breaklinks=true,bookmarks=false]{hyperref}
\else
\usepackage[pagebackref=true,breaklinks=true,colorlinks,bookmarks=false]{hyperref}
\fi

\ifwacvfinal
\else
\fi

\makeatletter
\newcommand{\printfnsymbol}[1]{%
  \textsuperscript{\@fnsymbol{#1}}%
}

\begin{document}

\title{Noisy Concurrent Training for Efficient Learning under Label Noise}

\author{Fahad Sarfraz\thanks{Equal contribution.Accepted as a conference paper at WACV 2021.}, Elahe Arani\printfnsymbol{1}, Bahram Zonooz\\
Advanced Research Lab, NavInfo Europe, Eindhoven, The Netherlands\\
{\tt\small \{fahad.sarfraz, elahe.arani, bahram.zonooz\}@navinfo.eu}
}

\maketitle


\begin{abstract}
Deep neural networks (DNNs) fail to learn effectively under label noise and have been shown to memorize random labels which affect their generalization performance.
We consider learning in isolation, using one-hot encoded labels as the sole source of supervision, and a lack of regularization to discourage memorization as the major shortcomings of the standard training procedure. 
Thus, we propose \textit{Noisy Concurrent Training (NCT)} which leverages collaborative learning to use the consensus between two models as an additional source of supervision.
Furthermore, inspired by trial-to-trial variability in the brain, we propose a counter-intuitive regularization technique, \textit{target variability}, 
which entails randomly changing the labels of a percentage of training samples in each batch as a deterrent to memorization and over-generalization in DNNs.
Target variability is applied independently to each model to keep them diverged and avoid the confirmation bias.
As DNNs tend to prioritize learning simple patterns first before memorizing the noisy labels, we employ a dynamic learning scheme whereby as the training progresses, the two models increasingly rely more on their consensus.
NCT also progressively increases the target variability to avoid memorization in later stages.
We demonstrate the effectiveness of our approach on both synthetic and real-world noisy benchmark datasets.
\end{abstract}

\section{Introduction}
Much of the recent advances in deep learning can be attributed to supervised learning algorithms which require huge amounts of annotated data~\cite{deng2009imagenet,lin2014microsoft}.
However, manually annotating the data is laborious and usually expensive task~\cite{neuhold2017mapillary} which can be prone to error when not verified by multiple annotators.
Furthermore, to utilize the widespread open-source data, various techniques were proposed for automatically annotating the data using user tags and keywords~\cite{makadia2008new,tsai2008automatically} and scaling up crowd-sourced datasets~\cite{mozafari2014scaling}.
While these approaches allow the creation of large datasets for training, they lead to noisy annotations.
A number of studies have shown that label noise has an adverse effect on the performance of the models~\cite{frenay2013classification,sukhbaatar2014training,zhang2016understanding}.
It is therefore pertinent to adapt the training procedure to leverage these datasets. 

\begin{figure}[tb]
 \centering
 \includegraphics[trim=0 0 0 0 clip, width=1\columnwidth]{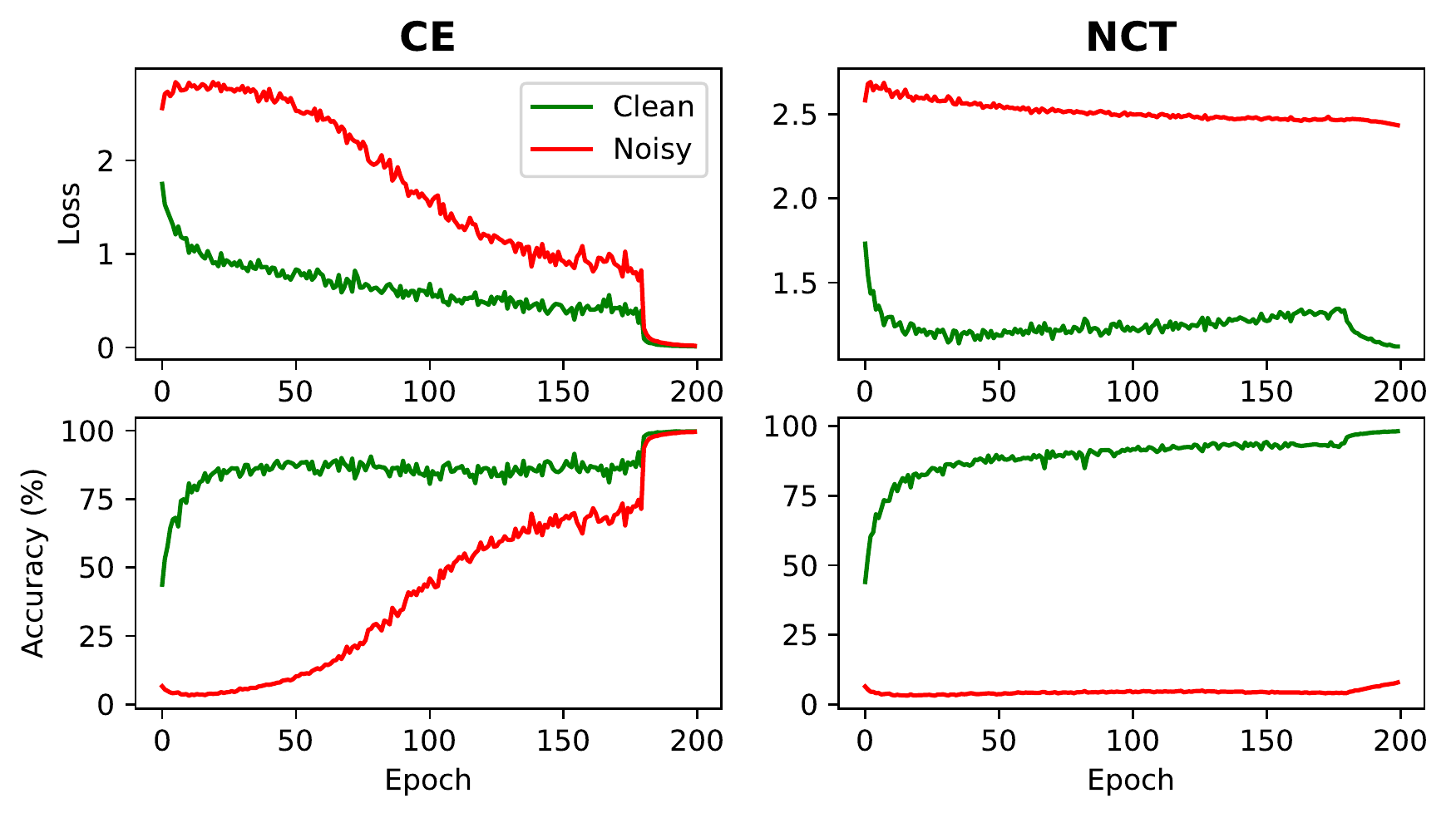}
 \caption{\footnotesize Average cross-entropy loss and accuracy on CIFAR-10 with 50\% symmetric label noise for the training samples with clean and noisy labels across the training epoch. Left: As training progresses standard model with cross-entropy loss (CE) memorizes the noisy labels. Right: Our proposed method, noisy concurrent training (NCT) effectively prevents the models from memorizing the noisy labels even though no distinction is made between them during training.}
 \label{fig:noisy-vs-clean}
\end{figure}

Deep neural networks (DNNs) have been shown to easily fit random labels~\cite{arpit2017closer} which makes it challenging to train the models efficiently.
The majority of the existing methods for training under label noise can be broadly categorized into two approaches: i) correcting the labels by estimating the noise transition matrix~\cite{goldberger2016training, patrini2016making}, ii) identifying the noisy labels to either filter out~\cite{han2018co, yu2019does} or down-weight those samples~\cite{jiang2017mentornet, malach2017decoupling}.
However, the former approach depends on accurately estimating the noise transition matrix which is difficult especially for a high number of classes, and the latter approach requires an efficient method for identifying noisy labels and/or an estimate of the percentage of noisy instances.
Amongst these, there has been more focus on separating the noisy and clean instances where a common criterion is to consider low-loss instances as a proxy for clean labels~\cite{arazo2019unsupervised,han2018co}.
However, harder instances can be perceived as noisy and hence the model can be biased towards easy instances. Both approaches consider the annotations quality as the primary reason for the decrease in model's performance and hence the proposed solutions rely on accurately relabeling, filtering out or down-weighting instances with incorrect labels.

Here we provide an alternative viewpoint on the issue of learning with noisy labels and attempt to improve the robustness of the underlying training framework.
We focus on the insufficiency of the standard training method. The cross-entropy loss maximizes a bound on the mutual information between one-hot encoded labels and the learned representation.
The model receives no information about the similarity of a data point among the classes and hence when the provided label is incorrect, it has no source of useful information about the instance or extra supervision to mitigate the adverse effect of the noisy label.
There is also a lack of regularization to discourage the model from memorizing the training labels.

To overcome these issues, we propose \textit{noisy concurrent training (NCT)} which introduces variability in supervision signal in a collaborative learning framework and takes advantage of building consensus among two different models.
Each model, in addition to a supervised learning loss, is trained with a mimicry loss that aligns the posterior distributions of the two models for building consensus on the secondary class probabilities as well as the primary class prediction.
To discourage memorization, we derive inspiration from neuroscience where the role of noise in the nervous system has been extensively studied.
Based on trial-to-trial response variation in the brain~\cite{scaglione2011trial} and the constructive role noise plays in forcing the biological neural networks to be more robust and explore more states~\cite{faisal2008noise}, we propose to use a counter-intuitive regularization technique we refer to as \textit{target variability} as a deterrent to memorization and over-generalization in DNNs.

Specifically, target variability entails randomly changing the labels of a percentage of training samples in a batch, independently for each model.
In addition to discouraging memorization, this keeps the two models sufficiently diverged and therefore retains the benefits of mutual learning, i.e. filtering different types of errors and avoiding confirmation bias in self-training.
Furthermore, since DNNs tend to learn simple patterns first and memorize the noisy labels in the later epochs~\cite{arpit2017closer}, NCT employs a dynamic learning scheme whereby as training progresses, the contribution of the supervised learning loss diminishes and the models focus more on building consensus. NCT also progressively increases the target variability to counter the higher tendency of DNNs to memorize the noisy labels at the later stages.
We show the efficacy of our proposed approach on noisy versions of CIFAR10, CIFAR100~\cite{krizhevsky2010cifar}, and Tiny-ImageNet~\cite{le2015tiny} as well as two real-world noisy datasets Clothing1M~\cite{xiao2015learning} and WebVision-v1~\cite{li2017webvision}.
Empirical results show the versatility and effectiveness of NCT under different noise types and noise levels.
In addition to improving the performance of the model on noisy datasets, NCT also improves the performance on clean datasets which demonstrates its utility as a general-purpose robust learning framework.

\begin{figure*}[t!h]
 \centering
 \includegraphics[trim=0 0 0 0 clip, width=.8\textwidth]{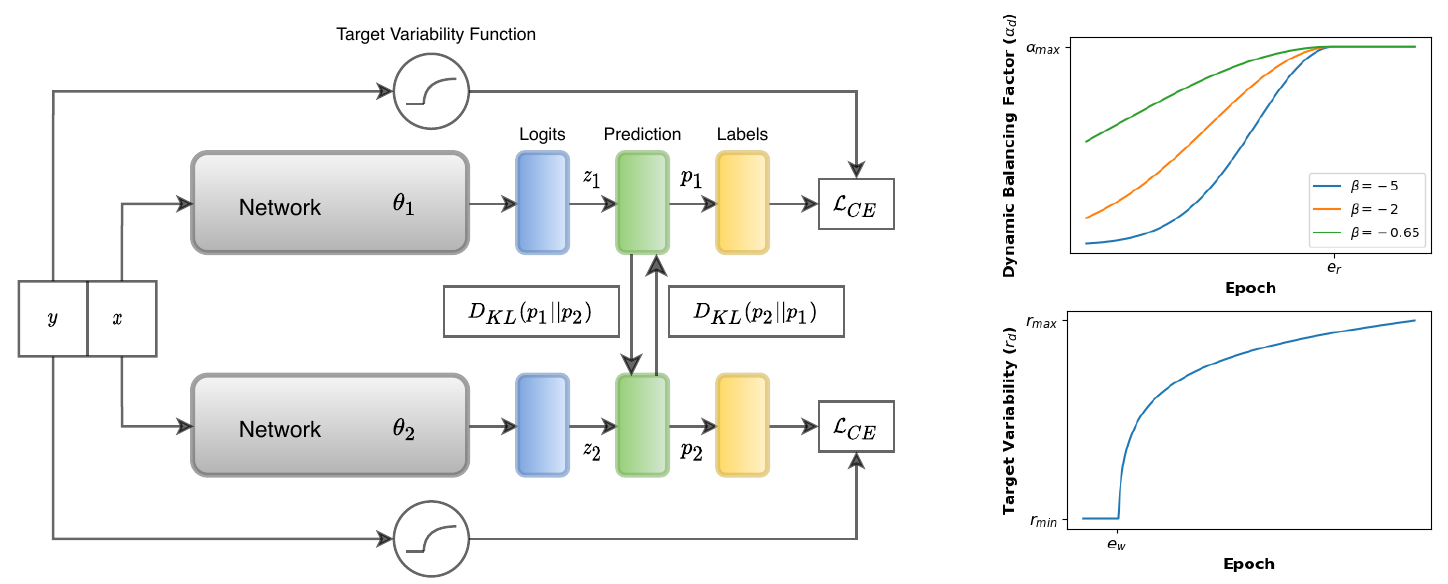}
 \caption{\footnotesize NCT involves training two models concurrently in a collaborative manner whereby each model is trained with a mimicry loss in addition to a supervised learning loss. The two models build consensus by aligning the posterior probabilities through the KL divergence loss. On each epoch, the (noisy) ground-truth labels for the two models are randomly flipped to a different class by a \textit{target variability function} with rate $r_d$. Target variability is applied independently for each model to prevent memorization and keep the two models diverged. NCT employs a dynamic learning scheme: (i)\textit{dynamic balancing function} progressively increases the weight of the mimicry loss $\alpha_d$ and (ii) \textit{target variability function} increases the target variability rate $r_d$ as training progresses.}
 \label{fig:NCT}
\end{figure*}

\section{Related Work}
The pervasiveness of label noise in real-world datasets has led to a number of approaches for training DNNs under noisy labels.
One approach is to implicitly or explicitly relabel the training labels. F-correction~\cite{patrini2016making} estimates the noise transition matrix to correct the noisy labels. However, estimating the noise transition matrix is particularly challenging for a higher number of classes. 
Joint-Optim~\cite{tanaka2018joint} uses a joint optimization framework of learning the model parameters and estimating true labels using the running average of the model's predictions. P-correction~\cite{yi2019probabilistic} models the labels as probability distributions over the classes and updates these distributions through back-propagation in an end-to-end manner.

Another approach involves correcting the loss function by reweighting the training samples.
Bootstrap~\cite{reed2014training} introduces a perceptual consistency term in the learning and uses a weighted combination of predicted and original labels as the correct labels. Instead of using a static weight for all samples, M-correction~\cite{arazo2019unsupervised} models sample loss with a beta mixture model to individually weigh each sample dynamically and adapts the Mixup~\cite{zhang2017mixup} augmentation. Mixup encourages the model to have linear behavior between samples and exhibits strong robustness to label noise. D2L~\cite{ma2018dimensionality} uses a similar combination of the original labels and network predictions depending on the dimensionality of the latent feature subspace.

A variant of the loss correction approach focuses on separating the clean and noisy instances. MentorNet~\cite{jiang2017mentornet} uses a predefined curriculum for selecting the clean instances but it is difficult to design a reliable criterion.
Decoupling~\cite{malach2017decoupling} trains two networks simultaneously and at each epoch only uses the training instances where the two models disagree for updating the models.
Disagreement amongst the two models, however, is not an optimal criterion for filtering out noisy labels and the disagreement region contains a number of noisy labels.
Co-teaching~\cite{han2018co} and Co-teaching+~\cite{yu2019does} use low loss training instances as a proxy for clean instances and use cross-update between the two models whereby each model selects the low-loss samples for the other model. These methods require an accurate estimate of the noise level which is difficult to obtain especially in the absence of a clean validation dataset.
Iterative-CV~\cite{chen2019understanding} randomly divides noisy datasets and utilizes cross-validation to identify clean samples before applying the co-teaching method on selected samples.
However, these approaches do not utilize noisy training instances for representation learning. Also using low-loss instances for identifying clean instances biases the model towards easy instances as hard instances are usually perceived as noisy.

There are a few other approaches such as Meta-Cleaner~\cite{zhang2019metacleaner} and Meta-Learning~\cite{li2019learning}. The former hallucinates clean representations of an object category according to a subset from the same category to identify noisy labels.
The latter proposes a gradient-based method to find model parameters that are more noise-tolerant.

The aforementioned approaches, in general, focus on accurately relabeling, filtering out, or down-weighting instances with incorrect labels as they consider the quality of the annotations as the primary reason for the model's failure to learn efficiently. Our proposed method, instead, focuses on improving the robustness of the underlying training framework.

\begin{algorithm*}[t]
\caption{Noisy Concurrent Training Algorithm}
\label{algo:nct}
\begin{algorithmic}[1]
\Statex {\bf Input:} {Dataset $D$, Number of classes $C$, Temperature $\tau$, Learning rate $\eta$, Batch size $b$, Total epochs $e_{max}$, Maximum target variability rate $r_{max}$, Warmup length $e_w$, Maximum alpha value $\alpha_{max}$, Ramp-up length $e_r$, Phase shift $\beta$}
\Statex \textbf{Initialize:} M1 and M2 parameterized by $\theta_1$ and $\theta_2$
\While{Not Converged}
\State {Sample a mini-batch: ${(x^{(1)}, y^{(1)}), ... , (x^{(b)},y^{(b)})} \sim D$}
\State Compute the dynamic balancing factor $\alpha_d$ based on Eq. \ref{eq:dynamic_alpha} 
\State Compute the target variability rate $r_d$ based on Eq. \ref{eq:target_variability}
\State Get the new targets: $\hat{y}_1, \hat{y}_2$ = \Call{Target\_Variability\_Function}{$\{y^{(1)},...,y^{(b)}\}$, $r_d$, C} (Algorithm \ref{algo:target-variability})
\State {Compute the loss functions for both M1 and M2 models:\par
$\mathcal{L}_{\theta_1} =(1-\alpha_d)\mathcal{L}_{CE}(\sigma(z_{\theta_2}), \hat{y}_1) + \alpha_d \tau^2 D_{KL}(\frac{\sigma(z_{\theta_1})}{\tau}||\frac{\sigma(z_{\theta_2})}{\tau})$ \par
$\mathcal{L}_{\theta_2} =(1-\alpha_d)\mathcal{L}_{CE}(\sigma(z_{\theta_2}), \hat{y}_2) + \alpha_d \tau^2 D_{KL}(\frac{\sigma(z_{\theta_1})}{\tau}||\frac{\sigma(z_{\theta_2})}{\tau})$}
\State {Compute stochastic gradients and update the parameters:\par
$\theta_1^*\gets\theta_1-\eta\frac{\partial\mathcal{L}_{\theta_1}}{\partial\theta_1}$\par
$\theta_2^*\gets\theta_2-\eta\frac{\partial\mathcal{L}_{\theta_2}}{\partial\theta_2}$}
\EndWhile
\Statex \Return{$\theta_1^*$ and $\theta_2^*$}
\end{algorithmic}
\end{algorithm*}

\section{Proposed Approach}
In this section, we first provide the motivation and intuition behind our method, Noisy Concurrent Training, and then formally present the different components of the proposed approach.

\subsection{Overview}
Our approach is loosely inspired by Boyd \etal~\cite{boyd2011cultural} study on cultural niche where they posit that the uniquely developed ability of humans to learn from others is absolutely crucial for human ecological success.
The authors suggest that cultural learning can increase the average fitness of the population only if it increases the ability of the population to create adaptive information.
A possible mechanism through which cultural learning can benefit the individual, as well as the population, is that it allows individuals to learn selectively - using environmental cues when they provide clear guidance and learning from others when they do not.
This ability to learn or imitate selectively is advantageous because opportunities to learn from experience or by observation of the world vary. 
Furthermore, some psychological models assume that our learning psychology has a genetically heritable information quality threshold that governs whether an individual relies on inferences from environmental cues or learns from others. Individuals with a low information quality threshold rely on even poor cues whereas individuals with a high threshold usually imitate. As the mean information quality threshold in the population increases, the fitness of learners increases because they are more likely to make accurate or low-cost inferences. At the same time, the frequency of imitators also increases.

Our proposed approach attempts to simulate the mechanism of cultural learning in neural networks. NCT involves training models concurrently whereby each model is trained with a convex combination of a supervised learning loss and a mimicry loss. Supervision from supervised loss can be considered as learning from the environmental cues whereas supervision from the mimicry loss can be viewed as imitation in cultural learning.
Even though the ground-truth labels (environmental cues) can be noisy, DNNs tend to prioritize learning simple patterns first before memorizing noisy labels, therefore in the initial phase of learning, the models can learn more from the supervised loss, gradually increasing the fitness of the two models (population). As training progresses, the information quality threshold can be increased and the model can rely more on imitating each other and building consensus. This is simulated using a dynamic balancing scheme which progressively increases the weight of the mimicry loss while reducing the weight of the supervised learning loss. This shifts the priority of the two models towards consensus building on their accumulated knowledge (model prediction) and aligning their posterior probability distributions. The mimicry loss provides an extra supervision signal for training the models in addition to the one-hot labels which can enable the models to learn useful information even from training samples with incorrect labels.

Furthermore, inspired by trial-to-trial variability in the brain, NCT employs a simple yet counter-intuitive regularization technique hereby referred to as \textit{Target Variability} whereby during training, the target labels of a fraction of samples are randomly changed for each batch independently for the two models.
Target variability serves multiple purposes: it implicitly increases the information quality threshold by indicating to the model that it cannot rely too much on the noisy labels,
acts as a strong deterrent to memorizing the training labels and also keeps the two models sufficiently diverged to avoid the confirmation bias arising from the method reducing to self-training. Figure \ref{fig:NCT} delineates the method.

\subsection{Formulation}
Given a dataset of $N$ samples, $D=\{(x^{(i)},{y}^{(i)})\}_{i=1}^N$, where $x^{(i)}$ is the input image and $y^{(i)} \in \{0, 1\}^C$ is the one-hot ground-truth label over $C$ classes which can be noisy, we formulate our proposed method, NCT, as dynamic collaboration learning between a cohort of two networks parametrized by $\theta_{1}$ and $\theta_{2}$.
Each network is trained with a supervised loss (standard cross-entropy, $\mathcal{L}_{CE}$) and a mimicry loss (Kullback–Leibler divergence, $D_{KL}$). The overall loss for each model is as follows:
\begin{equation}\label{eq:loss_m1}
\mathcal{L}_{\theta_1}= (1-\alpha)\mathcal{L}_{CE}(\sigma(z_{\theta_1}), y) + \alpha \tau^2 D_{KL}(\frac{\sigma(z_{\theta_2})}{\tau}||\frac{\sigma(z_{\theta_1})}{\tau})
\end{equation}
\begin{equation}\label{eq:loss_m2}
\mathcal{L}_{\theta_2} =(1-\alpha)\mathcal{L}_{CE}(\sigma(z_{\theta_2}), y) + \alpha \tau^2 D_{KL}(\frac{\sigma(z_{\theta_1})}{\tau}||\frac{\sigma(z_{\theta_2})}{\tau})
\end{equation}
where $\sigma$ is the softmax function, $z_{\theta}$ are the output logits and $\tau$ is the temperature which is usually set to 1. Using a higher $\tau$ value produces a softer probability distribution over classes. The balancing parameter $\alpha \in [0, 1]$ controls the relative weightage between the two losses.

For inference, we use the average ensemble of the two models,
\begin{equation}\label{eq:loss_m2}
y_{pred} = \sigma( \frac{z_{\theta_1} + z_{\theta_2}}{2})
\end{equation}

\subsection{Dynamic Balancing}
Given a mixture of clean and noisy labels, DNNs tend to prioritize learning simple patterns first and fit the clean data before memorizing the noisy labels~\cite{arpit2017closer}. NCT employs a dynamic balancing scheme whereby initially the two networks learn more from the supervision loss, i.e. smaller $\alpha_d$ value, and as the training progresses, the networks focus more on building consensus and aligning their posterior distribution through $D_{KL}$, i.e $\alpha_d \rightarrow 1$. To simulate this behavior, we use a sigmoid ramp-up function following~\cite{laine2016temporal},
\begin{equation}\label{eq:dynamic_alpha}
    \alpha_d = \alpha_{max}\exp{(-\beta(1-\frac{e}{e_{r}})^2})
\end{equation}
where $\alpha_{max}$ is the maximum alpha value, $e$ is the current epoch, $e_r$ is the ramp-up length (the epoch at which $\alpha_d$ reaches the maximum value) and $\beta$ controls the shape of the function. Figure \ref{fig:NCT} shows the dynamic balancing functions for different values of $\beta$.

\subsection{Dynamic Target Variability}
To mimic the trial-to-trial variability in the brain, variations in neural responses to the same stimuli, NCT uses \textit{target variability} whereby for each sample in the training batch, with probability $r$, the one-hot labels are changed to a random class sampled from a uniform distribution over the number of classes $C$. Target variability acts as a regularizer and discourages the model from memorizing the labels. Target variability is applied independently to each model so that the two networks remain sufficiently diverged so that collectively they can filter different types of errors. As the networks tend to memorize the noisy labels in later stages of training, NCT employs dynamic target variability whereby the target variability rate $r_d$ is lower for initial epochs and increases progressively during the training (Figure \ref{fig:NCT}). NCT uses a logarithmic ramp-up function,
\begin{equation}\label{eq:target_variability}
    r_d = 
    \begin{cases}
    r_{min} & \text{, if } e\leq e_{w}\\
    r_{min} + (r_{max} - r_{min}) \frac{\log[e - e_w]}{\log[e_{max} - e_w]} & \text{, otherwise}
    \end{cases}
\end{equation}
where $r_{min}$ and $r_{max}$ are the minimum and maximum target variability rates, $e$ is the current epoch, $e_{max}$ is the total number of epochs and $e_{w}$ is the warmup length. Figure \ref{fig:noisy-vs-clean} demonstrates the effectiveness of dynamic target variability in regularizing the model against memorizing the noise labels.
The details of the proposed method are summarized in Algorithms \ref{algo:nct} and \ref{algo:target-variability}.

\begin{algorithm}[t]
\caption{\textsc{Target\_Variability\_Function}}
\label{algo:target-variability}
\begin{algorithmic}[1]
\Statex {\bf Input:} {Labels $y$, mini-batch size $b$, Number of classes $C$, Target variability rate $r_d$}
\For {$i\in [1,2]$}
\State {Create the noise masks: \par
$m=[m_j \sim \mathcal{U}(0,1)]^b<r_d$}
\State {Sample the random targets:\par
$y_i=[l_j \sim \mathcal{U}(0,C-1) | l_j\neq y_j]^b$}
\State {Apply target variability and create the new targets: \par
$\hat{y}_i=m\odot y_i + (1-m)\odot y$}
\EndFor
\State \Return $\hat{y}_1$ and $\hat{y}_2$
\end{algorithmic}
\end{algorithm}

\section{Experimental Setup}
For our empirical analysis, we benchmark the performance of our approach on noisy versions of three different datasets CIFAR-10, CIFAR-100~\cite{krizhevsky2010cifar} and Tiny-ImageNet~\cite{le2015tiny} which represents classifications tasks of increasing complexity and are commonly used in literature to evaluate performance under noisy supervision~\cite{yu2019does,goldberger2016training}. 
We follow previous works~\cite{arazo2019unsupervised,li2019learning} on CIFAR-10 and CIFAR-100 where noise labels are generated by replacing a percentage of true labels with corrupted labels sampled uniformly from all the classes (i.e., the true label can be randomly maintained). For Tiny-ImageNet, we follow the experimental setup in~\cite{yu2019does} and test the performance of our model on two different types of label corruption: symmetry flipping and pair flipping. Here, symmetric noise is generated by replacing a percentage of true labels with corrupted labels sampled uniformly from the other classes (i.e., the true label cannot be maintained) whereas pair flipping simulates the scenario where annotators confuse between a pair of classes. 

It is important to note that the interplay of the hyperparameters of NCT is complementary in nature and therefore the desired effect can be achieved by keeping the majority of the parameters fixed and tuning only a few. For dynamic balancing, we fix $\tau=4$ and $\alpha_{max}=0.9$ as they are commonly used in knowledge distillation literature. In order to avoid overfitting to noisy labels in the initial training stage, we use $\beta=-0.65$ and $e_{r}$ is set to 90\% of the total epochs so that the transition of weight from supervised loss to mimicry is not too slow (Figure \ref{fig:NCT}, $\beta=-5$ vs $\beta=-0.65$). For dynamic target variability, we fix $r_{min}=0$ and $e_{w}=1$ while $r_{max}\in \{0, 0.1, 0.3, 0.5, 0.7, 0.9\}$ is selected using a small validation set. Hence, only the $r_{max}$ value is tuned for each experiment while the rest of the parameters remain constant.
Following \cite{arazo2019unsupervised}, we train our method on PreActResNet-18~\cite{he2016identity} and perform random crop and random horizontal flip followed by standard normalization. We train our models for 200 epochs using SGD with 0.9 momentum, weight decay of $1e$-$5$ and batch size 128. The initial learning rate of 0.02 is decayed by a factor of 10 after 180 epochs for CIFAR-10 and CIFAR-100 and 140 for Tiny-ImageNet. For CIFAR-10 we use $r_{max}$ values 0.1, 0.3, 0.5 for clean, symmetric-20 and symmetric-50, respectively. For CIFAR-100 we use $r_{max}=0.1$ for clean and $r_{max}=0.7$ for symmetric-20 and symmetric-50. For Tiny-ImageNet, we use $r_{max}=0.1$ for all the experiments. 

We further test the versatility of our method on two real-world noisy datasets Clothing1M~\cite{xiao2015learning} and WebVision-v1~\cite{li2017webvision}.
Clothing1M consists of 14 classes with one million training images collected from online shopping websites with auto-generated labels from surrounding text. Following previous works~\cite{arazo2019unsupervised, li2019learning}, we use ResNet-50 with ImageNet pretrained weights. We train the models for 200 epochs with an initial learning rate of 0.002 decayed by a factor of 10 at 180 epoch and $r_{max}=0.5$. For each epoch, we sample 1000 mini-batches of size 32 from the training data while ensuring the labels are balanced. WebVision contains 2.4 million images crawled from the Internet by using queries generated from the 1,000 semantic concepts of the benchmark ILSVRC 2012 dataset~\cite{deng2012large}. Following Chen \etal~\cite{chen2019understanding}, we train Inception-ResNet-v2~\cite{szegedy2016inception} models on the first 50 classes of the Google image subset. We train the models for 100 epochs with an initial learning rate of 0.01 decayed by a factor of 10 at 90 epoch and $r_{max}=0$. 

\begin{table*}[tb]
\centering
\caption{Comparison with prior methods on CIFAR-10 and CIFAR-100 datasets with symmetric noise. The results for baselines are copied from Arazo \etal~\cite{arazo2019unsupervised} and following them, we report the highest test accuracy (\%) across all epochs (Best) and the final epoch accuracy (Last). For our method, we report the average and 1 STD of three different seed values.}
\label{tab:cifar10-100}
\begin{tabular}{l|l|ccc||ccc}
\hline
 Dataset & \multicolumn{1}{c|}{} & \multicolumn{3}{c||}{CIFAR-10} & \multicolumn{3}{c}{CIFAR-100} \\\hline
Alg./Noise (\%) &  & 0 & 20 & \multicolumn{1}{c||}{50} & 0 & 20 & 50 \\ \hline
\multirow{2}{*}{Standard} & Best & 93.8 & 89.7 & 84.8 & 75.2 & 62.8 & 48.0 \\
 & Last & 93.7 & 81.8 & 55.9 & 75.1 & 62.7 & 40.8 \\ \hline
\multirow{2}{*}{Bootstrap~\cite{reed2014training}} & Best & 94.7 & 86.8 & 79.8 & 76.1 & 62.1 & 46.6 \\
 & Last & 94.6 & 82.9 & 58.4 & 75.9 & 62.0 & 37.9 \\\hline
\multirow{2}{*}{F-correction~\cite{patrini2016making}} & Best & 94.7 & 86.8 & 79.8 & 75.4 & 61.5 & 46.6 \\
 & Last & 94.6 & 83.1 & 59.4 & 75.2 & 61.4 & 37.3 \\\hline
\multirow{2}{*}{Mixup~\cite{zhang2017mixup}} & Best & 95.3 & \textbf{95.6} & 87.1 & 74.8 & 67.8 & 57.3 \\
 & Last & 95.2 & 92.3 & 77.6 & 74.4 & 66.0 & 46.6 \\\hline
\multirow{2}{*}{M-correction~\cite{arazo2019unsupervised}} & Best & 93.6 & 94.0 & \textbf{92.0} & 73.3 & 73.9 & \textbf{66.1} \\
 & Last & 93.4 & 93.8 & \textbf{91.9} & 71.3 & 73.4 & \textbf{65.4} \\\hline
\multirow{2}{*}{NCT} & \multirow{1}{*}{Best} & \textbf{95.6}$\pm$0.1 & 94.4$\pm$0.1 & 90.7$\pm$0.3 & \textbf{80.1}$\pm$0.1 & \textbf{74.4}$\pm$0.2 & 53.4$\pm$0.3 \\
 & \multirow{1}{*}{Last} & \textbf{95.5}$\pm$0.1 & \textbf{94.3}$\pm$0.0 & 89.7$\pm$0.3 & \textbf{80.0}$\pm$0.2 & \textbf{74.1}$\pm$0.1 & 52.3$\pm$0.7\\\hline
\end{tabular}
\end{table*}

\begin{table*}[tb]    
\centering
\caption{Comparison with prior methods on Tiny-ImageNet dataset with symmetric and asymmetric pair flip noise. The results for baselines are copied from Yu \etal~\cite{yu2019does} and following them, we report the highest (Best) and the average (Avg.) test accuracy (\%) over the last 10 epochs. For a fair comparison, we run M-Correction on the noise simulation in \cite{yu2019does} using their public code and hyperparameters mentioned in their paper. We also run Standard and Co-teaching+ on clean dataset. For all these experiments performed by us, we report the mean and 1 STD of three different seed values.}
\label{tab:tiny-imagenet}
\begin{tabular}{l|cc|cc|cc||cc}
\hline
Noise Type & \multicolumn{6}{c||}{Symmetric} & \multicolumn{2}{c}{Asymmetric} \\ \hline 
Noise (\%) & \multicolumn{2}{c|}{0} & \multicolumn{2}{c|}{20} & \multicolumn{2}{c||}{50} & \multicolumn{2}{c}{45} \\\hline
Alg. & Best & Avg. & Best & Avg. & Best & Avg. & Best & Avg. \\\hline
Standard & 57.4$\pm$0.5 & 56.7$\pm$0.5 & 35.8 & 35.6 & 19.8 & 19.6 & 26.32 & 26.2 \\
Decoupling~\cite{malach2017decoupling} & - & - & 37.0 & 36.3 & 22.8 & 22.6 & 26.61 & 26.1 \\
F-correction~\cite{patrini2016making} & - & - & 44.5 & 44.4 & 33.1 & 32.8 & 0.67 & 0.6 \\
MentorNet~\cite{jiang2017mentornet} & - & - & 45.7 & 45.5 & 35.8 & 35.5 & 26.61 & 26.2 \\
Co-teaching+~\cite{yu2019does} & 52.4$\pm$0.2 & 52.1$\pm$0.2 & 48.2 & 47.7 & 41.8 & 41.2 & 26.87 & 26.5 \\
M-correction~\cite{arazo2019unsupervised} & 57.7$\pm$0.3 & 57.2$\pm$0.4 & 57.2$\pm$0.5 & 56.6$\pm$0.4 & \textbf{51.6}$\pm$0.3 & \textbf{51.3}$\pm$0.3 & 24.8$\pm$10.0 & 24.1$\pm$10.3 \\
NCT & \textbf{62.4}$\pm$0.5 & \textbf{61.5}$\pm$0.2 & \textbf{58.0}$\pm$0.2 & \textbf{57.2}$\pm$0.3 & 47.8$\pm$0.1 & 47.4$\pm$0.2 & \textbf{43.0}$\pm$0.2 & \textbf{42.4}$\pm$0.1\\\hline
\end{tabular}
\end{table*}

\section{Results}
Here we first compare NCT with the priors works on both simulated noisy datasets and real-world noisy datasets, and then analyze the effect of the different components of the proposed method.

\subsection{Comparison with Prior Works}
We compare NCT with multiple baseline methods under similar experimental setup. Since the quality of the dataset is not known a priori, the learning method should be general to work in both noisy as well as clean datasets. For this reason, we compare our method on both clean and various levels of label noise. Table \ref{tab:cifar10-100} shows consistent improvement for lower noise levels. On clean CIFAR-100, the gap between M-Correction and NCT is considerable.
However, our method does not perform well compared to M-Correction for very high levels of symmetric noise (50\%). 
Table \ref{tab:tiny-imagenet} shows that the effectiveness of our approach generalizes beyond CIFAR datasets to the complicated Tiny-ImageNet classification task. On symmetric noise, we see a similar pattern as on CIFAR datasets. For asymmetric noise, which perhaps better simulates real-world noise, NCT provides a significant improvement in generalization. M-Correction shows an unstable behavior on asymmetric noise, indicated by the high standard deviation in performance.
Notably, there is considerable performance gap on clean dataset between NCT and other methods on the more challenging CIFAR-100 and Tiny-ImageNet datasets. This can be attributed to the fact that NCT does not make strong assumptions about label noise distribution or attempt to identify noisy labels. In the absence of an ideal separation criterion, clean samples particularly hard ones, can be wrongly identified as noisy samples and subsequently removing them or diminishing their influence can adversely affect performance. This effect is more pronounced as the number of classes increases.

To verify the practical usage of NCT, we also evaluate the method on two real-world noisy datasets. Table \ref{tab:webvision} shows that NCT provides a considerable performance gain ($\sim$10\% increase in top1 accuracy) over the prior methods on the WebVision dataset. For Clothing1M, Table \ref{tab:clothing1M} provides marginal gain over P-correction.

\begin{table}[tb]
\centering
\caption{Comparison with prior methods trained on WebVision dataset. The results for baselines are copied from Chen \etal~\cite{chen2019understanding} and following them, we report the final accuracy (\%) on the WebVision and ImageNet ILSVRC12 validation sets. For our method, we report the mean and 1 STD of three different seed values.}
\label{tab:webvision}
\begin{tabular}{l|cc|cc}
\hline
Alg./Dataset & \multicolumn{2}{c|}{WebVision} & \multicolumn{2}{c}{ILSVRC12} \\ \hline
 & top1 & top5 & top1 & top5 \\\hline
F-correction~\cite{patrini2016making} & 61.12 & 82.68 & 57.36 & 82.36 \\
Decoupling~\cite{malach2017decoupling} & 62.54 & 84.74 & 58.26 & 82.26 \\
D2L~\cite{ma2018dimensionality} & 62.68 & 84.00 & 57.80 & 81.36 \\
MentorNet~\cite{jiang2017mentornet} & 63.00 & 81.40 & 57.80 & 79.92 \\
Co-teaching~\cite{han2018co} & 63.58 & 85.20 & 61.48 & 84.70 \\
Iterative-CV~\cite{chen2019understanding} & 65.24 & 85.34 & 61.60 & 84.98 \\
\hline
\multirow{2}{*}{NCT} & \bf 75.16 & \bf 90.77 & \bf 71.73 & \bf 91.61\\
& $\pm$0.34 & $\pm$0.27 & $\pm$0.44 & $\pm$0.22\\\hline
\end{tabular}
\end{table}

\begin{table}[tb]
\centering
\caption{Comparison with prior methods on Clothing1M. The results for baselines are copied from original papers and following them, we report the best test accuracy (\%). For our method, we report the mean and 1 STD of three different seed values.}
\label{tab:clothing1M}
\begin{tabular}{l|c}
\hline
Alg. & \multicolumn{1}{c}{Test Accuracy} \\ \hline
Standard & 68.94 \\
F-correction~\cite{patrini2016making} & 69.84 \\
Joint-Optim~\cite{tanaka2018joint} & 72.16 \\
M-correction~\cite{arazo2019unsupervised} & 71.00 \\
Meta-Cleaner~\cite{zhang2019metacleaner} & 72.50 \\
Meta-Learning~\cite{li2019learning} & 73.47 \\
P-correction~\cite{yi2019probabilistic} & 73.49 \\
NCT & \textbf{74.02}$\pm$0.08 \\\hline
\end{tabular}
\end{table}

The empirical results on both clean and noisy versions of benchmark datasets as well as consistent improvement on real-world noisy datasets demonstrate the effectiveness of NCT as a general-purpose learning framework that is robust to label noise. Our method does not perform well on very high levels of noise, as it does not involve identifying clean and noisy samples and treating them differently as the goal of the study is to improve the noise tolerance of the underlying training framework. However, we argue that perhaps very high levels of symmetric noise, e.g. 50\% or 90\%, do not truly represent the nature of label noise in real-world datasets. While we can expect a considerable amount of label noise, greater than or close to 50\% would be implausible. Also, real-world datasets mostly contain structured (asymmetrical noise) with confusion between visually similar classes.

Furthermore, harder samples, where the orientation of the object, size, position, or the background makes it less indistinguishable from other classes, are more likely to be misclassified rather than all data points within an object class having an equal chance of being incorrectly labeled. Though still not truly representative, perhaps asymmetric pair flip noise is closer to noise distributions in the real world. Therefore, while these synthetic noisy datasets provide us with key insights and help in comparing the utility of various approaches, overemphasis on high levels of synthetic noise can potentially bias our methods towards noise distributions that are not representative of the real-world noisy datasets. This is particularly applicable to methods which focus on identifying the noisy labels where the noise distribution plays a more crucial role. We hope to bring into attention the need for a uniform set of synthetic noise distributions which are more representative of real-world label noise distributions to better study the characteristics of these datasets and benchmark the utility of different methods. In addition to these, real-world noisy datasets can provide a better estimate of the utility of the proposed approaches in the practical setting.

\subsection{Information Compression}
\begin{figure}[tb]
 \centering
 \includegraphics[trim=0 0 0 0 clip, width=0.9\columnwidth]{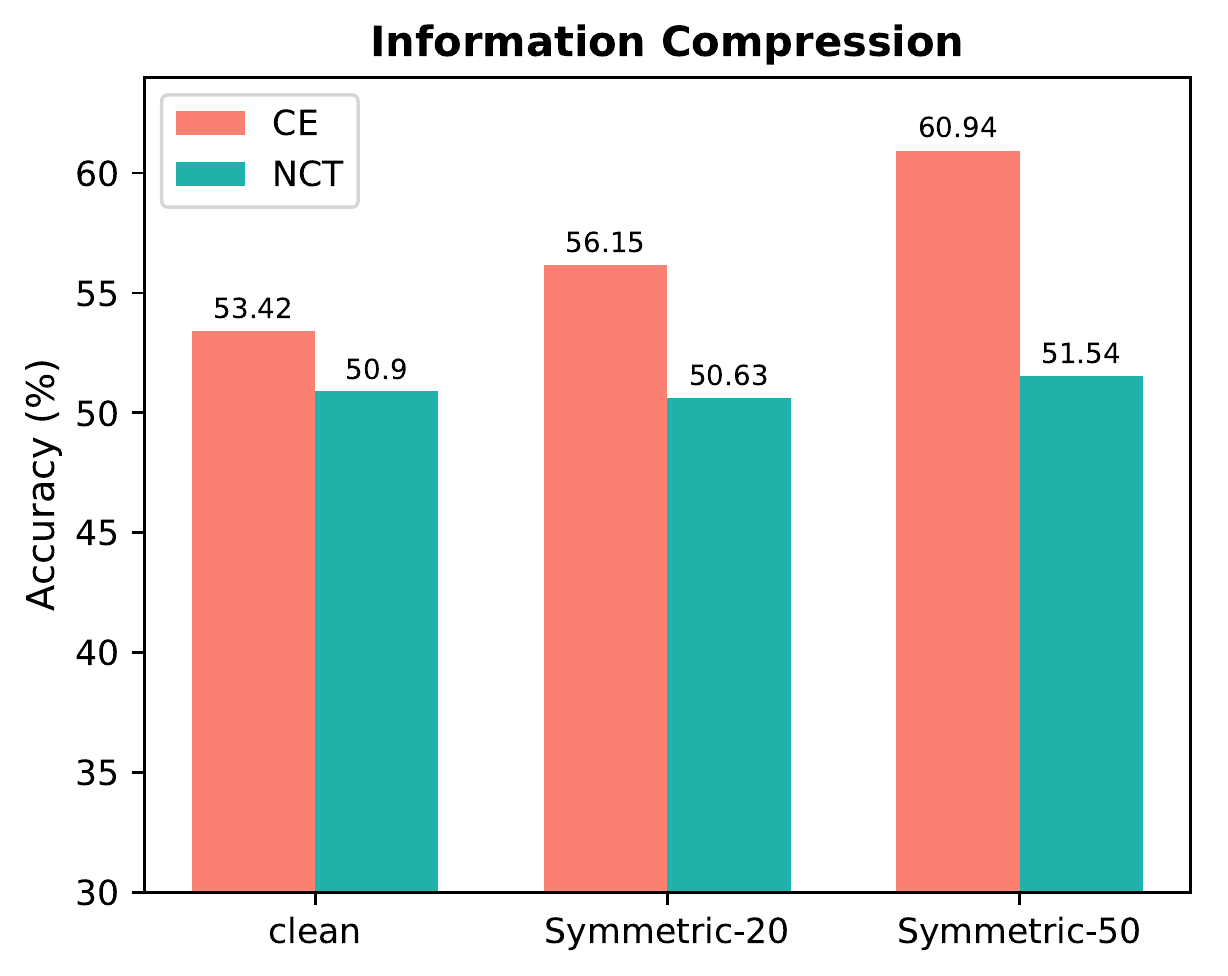}
 \caption{\footnotesize Comparative analysis on the degree to which standard training (CE) and Noisy Concurrent Training (NCT) with frozen learned representations (under varying levels of label noise) can fit binary random labels. Lower training error indicates higher information compression.}
 \label{fig:info_comp}
\end{figure}

To study the effect of our proposed method on the learned representations of the models, we follow the analysis in \cite{lamb2019interpolated} to do a comparative study on the effectiveness of NCT to compress information in learned representations relative to standard training under noisy labels.
A number of studies~\cite{tishby2015deep,shwartz2017opening} have shown that there is a relationship between the compression of information in the features learned by DNNs and their generalization. They relate the degree of information compression in the network's hidden states to bounds on generalization whereby stronger compression leads to strong generalization bounds.
To this end, we freeze the learned representations of the model and study how well the frozen representations can fit random binary labels. For NCT, we pick one of the two trained models.
We add a 2-layer multi-layer perceptron (MLP) network with 400 and 200 neurons on top of the frozen representations of PreActResNet-18 model trained on CIFAR-10 under varying degrees of symmetric label noise and fit them on random binary labels.
For all experiments, we select the first two classes and assign random labels i.e. the model fits 10000 random labels.
The difficulty in fitting the random variables show how well the model compresses information in the learned representations. Therefore, lower training accuracy shows better information compression.
Figure \ref{fig:info_comp} shows that NCT is able to consistently learn more compressed features compared to standard training. 

\begin{table}[tb]
\centering
\caption{Effect of target variability rate parameter, $r_{max}$, on CIFAR-10. We report the highest test accuracy (\%) across all epochs (Best) and the final epoch accuracy (Last). The mean and 1 STD of three different seed values are reported.}
\label{tab:r_max}
\begin{tabular}{l|l|c|c}
\hline
\multicolumn{2}{l|}{} & \multicolumn{2}{c}{Symmetric (\%)} \\ \cline{3-4} 
\multicolumn{2}{l|}{$r_{max}$} & 20 & 50 \\\hline
\multirow{2}{*}{0.0} & Best & 94.25$\pm$0.12 & 85.37$\pm$0.27 \\
 & Last & 93.94$\pm$0.15 & 79.60$\pm$0.17 \\\hline
\multirow{2}{*}{0.1} & Best & 94.26$\pm$0.09 & 86.56$\pm$0.20 \\
 & Last & 94.08$\pm$0.08 & 81.00$\pm$0.23 \\\hline
\multirow{2}{*}{0.3} & Best & {\bf 94.40}$\pm$0.07 & 89.35$\pm$0.29 \\
 & Last & {\bf 94.25}$\pm$0.03 & 86.83$\pm$0.32 \\\hline
\multirow{2}{*}{0.5} & Best & 94.25$\pm$0.12 & {\bf 90.70}$\pm$0.28 \\
 & Last & 94.19$\pm$0.09 & {\bf 89.74}$\pm$0.29 \\\hline
\multirow{2}{*}{0.7} & Best & 93.33$\pm$0.08 & 89.69$\pm$0.07 \\
 & Last & 93.21$\pm$0.02 & 89.48$\pm$0.25 \\\hline
\multirow{2}{*}{0.9} & Best & 88.20$\pm$0.24 & 82.88$\pm$0.36 \\
 & Last & 87.05$\pm$0.13 & 72.23$\pm$0.27\\\hline
\end{tabular}
\end{table}

\subsection{Effect of Target Variability}
Here, we analyze the sensitivity of our method to the target variability parameters. We use the CIFAR-10 dataset with the same experimental setup as for our previous experiments and show the effect of changing the $r_{max}$ value while keeping all other parameters fixed. Table \ref{tab:r_max} shows that target variability provides significant performance gain compared to the baseline NCT method without target variability ($r_{max} = 0$). Generally, for a wide range of target variability rates, $0.3 \leq r_{max} \leq 0.7$, NCT is not very sensitive to the choice of $r_{max}$ value. The method is more sensitive to the $r_{max}$ value for higher noise levels (50\%) compared to the lower noise levels (20\%).

\subsection{Ablation Study}
To analyze the effect of individual components of NCT, we sequentially remove components from the final method and see how the performance is affected. We use the Tiny-ImageNet dataset with the same experimental setup as our previous experiments.
(a) The performance drop with NCT w/o EN, where only one model, $\theta_{1}$ is used for inference at test time while the training process remains unchanged, show that the ensemble of two diverged models in NCT consistently provides improvement in performance.
(b) The effect of removing target variability from NCT, NCT w/o TV, is more pronounced for higher noise level. For Symmetric 50 and Asymmetric 45, target variability provides considerable gain while it marginally reduces for Symmetric 20. 
(c) Removing the target variability and dynamic balancing (NCT w/o (TV + DB)) reduces NCT to Deep Mutual Learning (DML)~\cite{zhang2018deep} which replaces the one-way knowledge transfer from a large pretrained model in traditional knowledge distillation~\cite{hinton2015distilling} with knowledge sharing between a cohort of compact models trained collaboratively. The significant drop suggests that progressively shifting the focus of learning from the training labels to building consensus increases the effectiveness of the method to learn under label noise.
(d) Finally, the gap between Standard and DML across all noise variations show the effectiveness of collaborative learning under label noise.
This shows that all the components contribute to the robustness of NCT.

\begin{table}[tb]
\centering
\caption{Ablation study on the Tiny-ImageNet dataset. we report the highest (Best) and the average (Avg.) test accuracy (\%) over the last 10 epochs. The mean and 1 STD of three different seed values are reported. EN, TV and DB stand for ensemble inference, target variability, and dynamic balancing, respectively.}
\label{tab:ABLATION}
\resizebox{\columnwidth}{!}{%
\begin{tabular}{l|l|cc|c}
\hline
 \multicolumn{2}{l|}{Noise Type} & \multicolumn{2}{c|}{Symmetric} & Asymmetric \\\hline
 \multicolumn{2}{l|}{Noise (\%)} & 20 & 50 & 45 \\\hline
\multirow{2}{*}{NCT} & Best & 58.0$\pm$0.2 & \textbf{47.8$\pm$0.1} & \textbf{43.0$\pm$0.2} \\
 & Avg. & 57.2$\pm$0.3 & \textbf{47.4$\pm$0.2} & \textbf{42.4$\pm$0.1}\\\hline
 \multirow{2}{*}{NCT w/o EN} & Best & 57.0$\pm$0.4 & 46.8$\pm$0.1 & 42.5$\pm$0.3 \\
 & Avg. & 56.2$\pm$0.2 & 46.3$\pm$0.2 & 41.6$\pm$0.1\\\hline
\multirow{2}{*}{NCT w/o TV} & Best & \textbf{58.1$\pm$0.3} & 47.0$\pm$0.2 & 42.2$\pm$0.3 \\
 & Avg. & \textbf{57.6$\pm$0.3} & 46.4$\pm$0.2 & 41.5$\pm$0.3 \\\hline
\multirow{2}{*}{NCT w/o (TV + DB)} & Best & 54.0$\pm$0.4 & 40.0$\pm$0.3 & 39.2$\pm$0.4 \\
 & Avg. & 53.1$\pm$0.4 & 39.2$\pm$0.3 & 38.3$\pm$0.4 \\\hline
\multirow{2}{*}{Standard} & Best & 42.1$\pm$0.3 & 24.1$\pm$0.3 & 31.4$\pm$0.5 \\
 & Avg. & 41.1$\pm$0.1 & 23.2$\pm$0.2 & 30.2$\pm$0.2 \\\hline
\end{tabular}}
\end{table}

\section{Conclusion}
In this paper, we proposed Noisy Concurrent Training which involves training a cohort of two models in conjunction and building consensus among the two models in addition to the supervised learning loss. The method dynamically shifts the focus of learning from fitting the training labels in the initial learning phases towards building consensus in the later stages. The method also employs target variability as deterrent to memorization and progressively increases the variability during training. We showed the effectiveness of our method on multiple synthetic noisy datasets with varying degrees and types of label noise as well as real-world noisy datasets. Our study shows that increasing the robustness of the underlying training framework as an alternative to filtering and down-weighting noisy labels is a promising direction.
{\small
\bibliographystyle{ieee_fullname}
\bibliography{egbib}
}

\end{document}